\documentclass{article}

     \PassOptionsToPackage{numbers, compress}{natbib}

     \usepackage[preprint]{neurips_2021}

\usepackage[utf8]{inputenc} 
\usepackage[T1]{fontenc}    
\usepackage{hyperref}       
\usepackage{url}            
\usepackage{booktabs}       
\usepackage{amsfonts}       
\usepackage{nicefrac}       
\usepackage{microtype}      
\usepackage{xcolor}         
\usepackage{graphicx}       
\usepackage{floatrow}       
\usepackage{wrapfig}        
\usepackage{multirow}       

\title{ Bootstrap Representation Learning for Segmentation on Medical Volumes and Sequences}

\author{%
	Zejian Chen\footnotemark[1] \\
	Shenzhen University\\
	\texttt{chenzejian19@email.szu.edu.cn} \\
	 \And
	 Wei Zhuo\footnotemark[1] \\
	 Tencent \\
	 \texttt{wei.zhuowx@gmail.com} \\
	 \AND
	 Tianfu Wang \\
	 Shenzhen University \\
	 \texttt{tfwang@szu.edu.cn} \\
	 \And
	 Wufeng Xue\footnotemark[2]\\
	 Shenzhen University \\
	 \texttt{xwolfs@hotmail.com} \\
	 \And
	 Dong Ni\footnotemark[2] \\
	 Shenzhen University \\
	 \texttt{nidong@szu.edu.cn} \\
}

\begin{document}
    \footnotetext[1]{Both authors contributed equally.}
    \footnotetext[2]{Corresponding author.}	 	
	
	\maketitle
	
	\begin{abstract}
    In this work, we propose a novel straightforward method for medical volume and sequence segmentation with limited annotations. To avert laborious annotating, the recent success of self-supervised learning(SSL) motivates the pre-training on unlabeled data. Despite its success, it is still challenging to adapt typical SSL methods to volume/sequence segmentation, due to their lack of mining on local semantic discrimination and rare exploitation on volume and sequence structures.  
	Based on the continuity between slices/frames and the common spatial layout of organs across volumes/sequences, we introduced a novel bootstrap self-supervised representation learning method by leveraging the predictable possibility of neighboring slices. At the core of our method is a simple and straightforward \textit{dense self-supervision on the predictions of local representations} and a strategy of \textit{predicting locals based on global context}, which enables stable and reliable supervision for both global and local representation mining among volumes. Specifically, we first proposed an asymmetric network with an attention-guided predictor to enforce distance-specific prediction and supervision on slices within and across volumes/sequences. Secondly, we introduced a novel prototype-based foreground-background calibration module to enhance representation consistency. The two parts are trained jointly on labeled and unlabeled data. When evaluated on three benchmark datasets of medical volumes and sequences, our model outperforms existing methods with a large margin of 4.5\% DSC on ACDC, 1.7\% on Prostate, and 2.3\% on CAMUS. Intensive evaluations reveals the effectiveness and superiority of our method. 
	
	\end{abstract}
	
	\section{Introduction}
	Segmentation for medical volume and sequences plays important role in clinical practice, including disease diagnosis, anatomical quantification, radiation therapy, treatment planning, and population studies~\cite{pham2000current, petersen2013imaging, zaidi2010pet, sharma2010automated}. Deep learning-based segmentation methods have achieved great success attribute to the massive labeled samples~\cite{ronneberger2015u, milletari2016v, zhou2018unet++}.
	However, collecting a large amount of annotations is usually impractical due to the considerable request of clinical expertise and time, especially for medical volumes and sequences. 
	
	To alleviate the dependencies on massive annotations, semi-supervised strategies apply joint training on labeled and unlabeled images with consistency regularization~\cite{tarvainen2017mean, miyato2018virtual, cui2019semi, berthelot2019mixmatch, sohn2020fixmatch}, and entropy minimization~\cite{grandvalet2005semi, lee2013pseudo, bai2017semi}.
	Given the recent success of self-supervision Learning (SSL)~\cite{chen2020simple, chen2020big, he2020momentum, grill2020bootstrap, chen2020exploring, caron2020unsupervised}, pre-training on unlabeled medical images is promising to provide a much better initialization for the subsequent fine-tuning. 
	SSL is a stream of methods that design pre-text training signals, such as rotation, colorization and jigsaw, based on the inherent attribute of training data to enable pre-training on unlabeled images. 
    Current SSL methods however mainly optimize on image-level objectives, and remains to be sub-optimal to the downstream task of dense predictions like segmentation. \cite{chaitanya2020contrastive} modifies the contrast learning~\cite{chen2020simple} for volume segmentation to leverage the inherent consistency and variation among subjects and provides the recent state-of-the-art. 
    However, the contrast is not always reasonable given the spatial continuity between neighboring slices and as well neighboring patches within a slice.  
    To avoid modeling contrast pairs in this scenario, we build our model based on the inherent predictable potentials between slices/frames in medical volume and sequences. 
	
	\begin{figure}[t]
		\centering
		\includegraphics[width=1.0\textwidth, trim=0 100 0 60, clip]{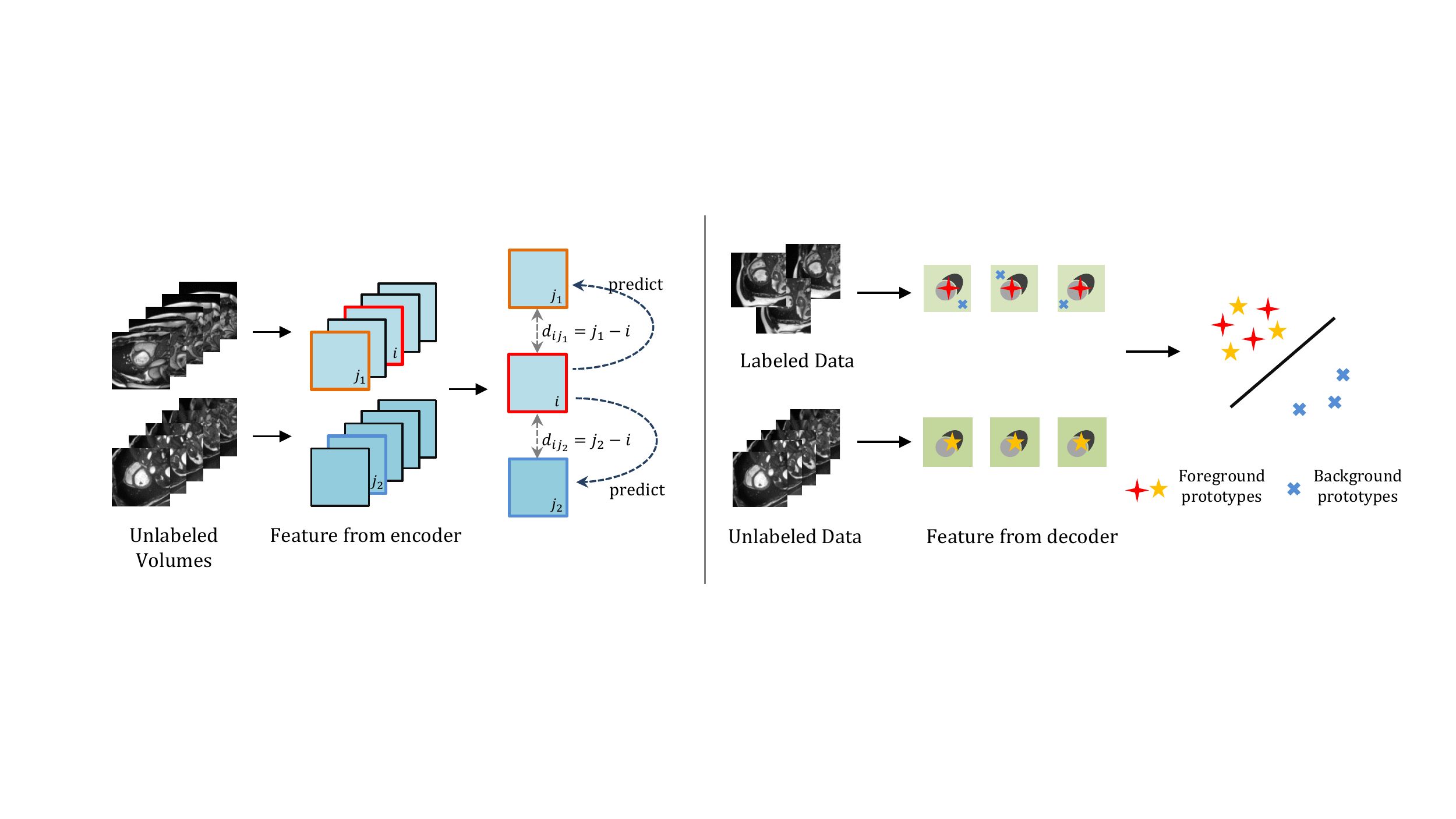}
		\caption{Conceptual illustration of our bootstrap representation method. Left: Slice prediction across and within volumes in the representation space. Right: Foreground-background calibration for unlabeled data with reference local feature of labeled data.}
		\label{fig_method}
	\end{figure}
	
	In this work, we propose a novel bootstrap representation learning method for volume and sequence segmentation with limited annotations. 
    We introduce a stable and reliable self-supervision mechanism that can exploit both global and local structures of the medical volume and sequences, across- and within-subjects. 
    Because of the continuity of anatomical structures within one volume and the common spatial layout of the organ across different subjects, the change between neighboring slices is smooth and follows a specific pattern across different subjects. 
    In view of this, we assume that one slice can be used to predict other slices of the same volume and, to be ambitious, to predict slices from different volumes (see left of Fig.~\ref{fig_method}). Inspired by the SSL methods~\cite{grill2020bootstrap,chen2020exploring},
    we fully exploit the predictable possibility across- and within-subjects, and introduce a novel asymmetric architecture to predict from the current slice the dense feature maps of any rest slice based on their global context. Dense similarity loss is utilized to enable self-supervised training. \textit{Our simple dense prediction from global context and our dense similarity supervision} enables representation mining on both local and global levels. To make the predictor more flexible and reliable, a distance-specific attention-guided predictor is designed for our intra- and inter-subject slice prediction. 
	
    In addition, we introduce a novel foreground-background calibration module, which can calibrate for the features maps of the unlabeled data in the decoder, with reference features of the labeled data, see right of Fig.~\ref{fig_method}. We assume the target organ of a subject should be very different from the background of any slices of all the subjects. To this end, we propose a novel prototype-based semantic-aware contrastive loss for the calibration. The semantic information can be indicated by the foreground mask for the labeled data or the regions with maximal response in the probability map for the unlabeled data. Besides, the prototype-based contrasting is more stable than contrasting with local patches that have no definite semantics. 

    In summary, our work has the following contributions: 
	\begin{itemize}
	    \item  We design a novel joint training framework that enables training on labeled and unlabeled data at the same time. It can avoid knowledge forgotten on unlabelled data, which could happen in two-stage training~\cite{chaitanya2020contrastive}.
	    \item  We build a novel bootstrap strategy with an attention-guided predictor with distance reference to exploit the predictable possibility of dense representations within medical volume/sequence to enable effective representation learning from the unlabeled data.
	    \item We introduce a novel foreground-background calibration module based on prototype-based contrastive loss for stable local-region supervision. 
	    \item Intensive experiments with two volume datasets and one sequential dataset demonstrate the effectiveness and superiority of our method on segmentation of medical images. 
	\end{itemize}
	
	\section{Related works}
	
    \paragraph{Segmentation methods on medical volume and sequence}
	For segmentation on volume and sequence data, leveraging the 3D information has been proved an effective way to success, such as temporal consistency~\cite{pedrosa2017fast, qin2018joint, yan2018left, wei2020temporal} or spatial continuity~\cite{zhou2018semi, zhang2020inter}. To extract the spatial-temporal features, \cite{li2019recurrent} utilized an RNN-based model to aggregate the information. The optical flow between consecutive frames was estimated to collect the motion information~\cite{qin2018joint, yan2018left} or to propagate the labels~\cite{pedrosa2017fast}. \cite{wei2020temporal} leveraged temporal information by co-learning of segmentation and tracking in appearance and shape level. \cite{zhang2020inter} boosted the segmentation accuracy by learning the inter-slice spatial context explicitly. Despite their success, these methods rely on all accessible labeled data for training and may fail to obtain good results when only a few labeled cases are available. 
	
	\paragraph{Semi-supervised methods} 
    To alleviate the pressure of annotation, semi-supervised learning dedicates to learn from both unlabeled data and labeled data. Typical techniques in semi-supervised learning include consistency regularization~\cite{tarvainen2017mean, miyato2018virtual, cui2019semi, sohn2020fixmatch, berthelot2019mixmatch} and entropy minimization~\cite{grandvalet2005semi, lee2013pseudo, bai2017semi}. The former usually constrains the representation of unlabeled data under various perturbations to be similar, while the latter is gained in a self-training manner with pseudo-labeling. 
    In semantic segmentation, existing techniques in semi-supervised learning is widely used~\cite{bai2017semi, zhang2017deep, nie2018asdnet, cui2019semi, yu2019uncertainty, zou2020pseudoseg, xia20203d, chaitanya2021semi}. 
    Early works proposed iterative self-training approaches to generate pseudo labels by learning a good teacher model~\cite{bai2017semi} or utilizing a GAN-based model~\cite{zhang2017deep}. Recent works enforce the predictions to be consistent, either from augmented input images~\cite{zou2020pseudoseg, xia20203d} or confidence map~\cite{nie2018asdnet, yu2019uncertainty}. 
    Our work exploits an orthogonal direction based on a joint strategy of self-supervised representation learning and semi-supervised foreground/background calibration for representation mining. 
     
	\paragraph{Self-supervised learning}
	Self-supervised learning (SSL) in recent years gains great interest due to the massive realistic requirement of pre-training on unlabeled data. Typical SSL methods include pseudo-labeling~\cite{lee2013pseudo, bai2017semi}, deep clustering~\cite{caron2020unsupervised, caron2018deep}, contrast learning~\cite{he2020momentum, chen2020simple}, and those based on hand-craft objectives~\cite{komodakis2018unsupervised,noroozi2016unsupervised,pathak2016context,larsson2017colorization}. 
	In the context of medical image, many works learned the representation by leveraging anatomical position~\cite{bai2019self}, structure contexts on image~\cite{chen2019self} or volumes~\cite{zhuang2019self, zhu2020rubik, tao2020revisiting}. 
	These methods however are designed based on image-level semantic consistency and lack explicit exploitation on the representation of finer scales, which is important for segmentation. To enhance spatial sensitivity of local representations, \cite{xie2020propagate} proposed two pixel-level pretext tasks, one based on contrast learning, the other one based on pixel-to-propagation consistency. The latter one encourages representation consistency of corresponding pixels from sub-crops of the augmented views.
 	CGL~\cite{chaitanya2020contrastive} modeled the contrasts on both global and local patches, where patches of the corresponding locations from different subjects are positives and all others are negatives. Due to the fact that patch similarity is only based on local texture, it is hard to build reliable contrast relations.
 	
	Another recent research stream of SSL~\cite{grill2020bootstrap, chen2020exploring} prevent using negatives and attempt to predict one view of an instance from a different view via an asymmetric network with an additional predictor and the stop-gradient mechanism. We assume this is more straightforward and save efforts on parsing contrast relations. Even though, most of them work on global representation prediction and rarely exploit mid-/local-level regions. Our work builds on this stream and proposes a new framework for reliable mining on both global and local representations. In addition, different from the two-stage training in SSL, we adopt joint learning to avoid forgotten of the pre-training knowledge. 

	\section{Methods}
	
	\begin{figure}[t]
		\centering
		\includegraphics[width=\textwidth, trim=0 60 0 50, clip]{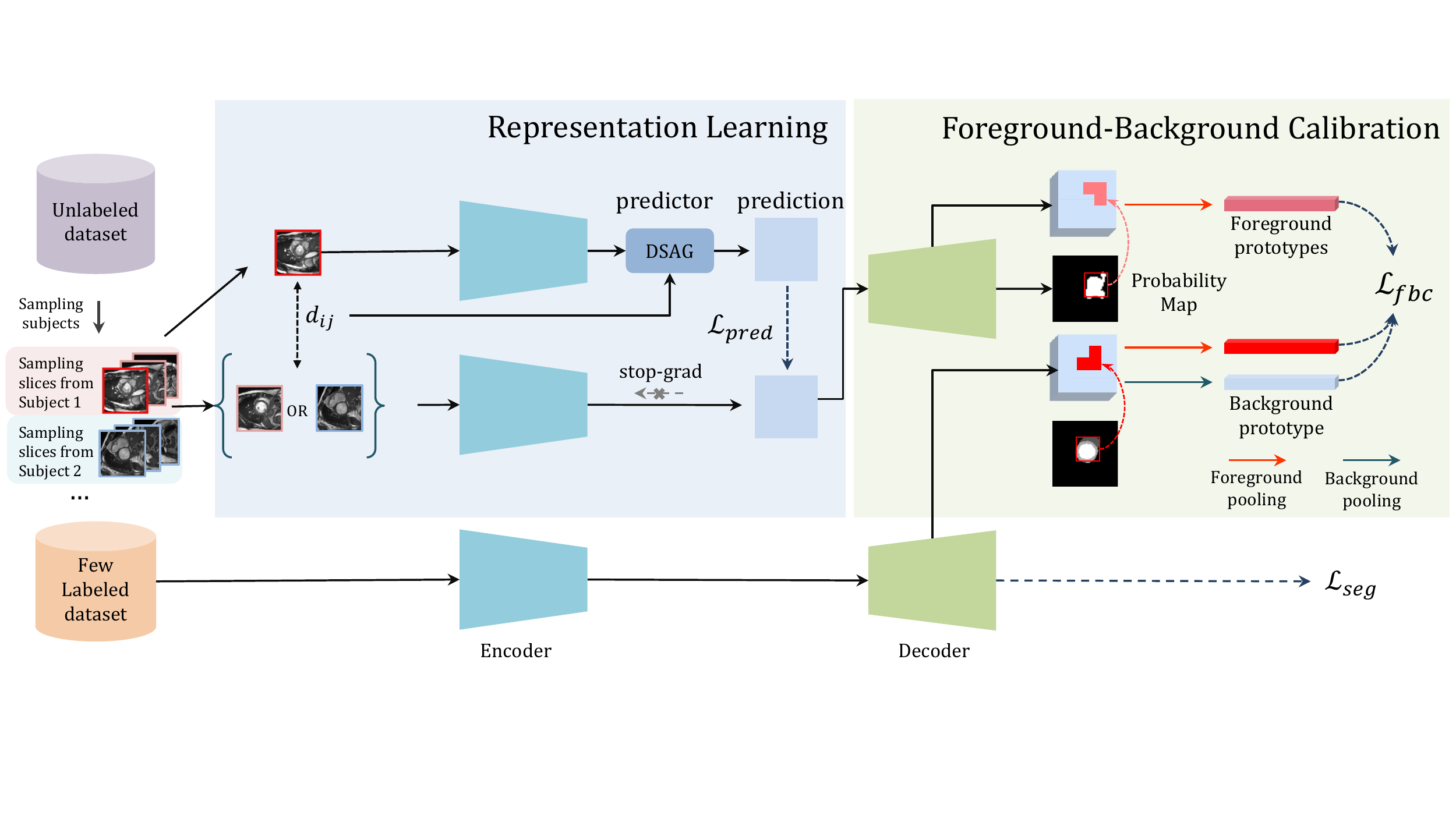}
		\caption{
		Network architecture of our method. Three modules are included: the bottom one is an encoder-decoder network for segmentation and trained with a few labeled data, the top-left one is the slice prediction module for representation mining with unlabeled data, and the top-right one is the foreground-background calibration module for the features from decoder.}
		\label{fig_pipeline}
	\end{figure}
	Suppose we have a labeled 3D volumetric medical segmentation dataset $D_{L} = \{(X^i, Y^i)\}, i\in \{1...N\}$, where $X^i$ consists a series of 2D slices $\{x_k^i\},k\in\{1...K\}$ for subject $i$ and $Y^i$ is the corresponding annotations $\{y_k^i\}$ for each slice. $ D_U = \{X^j\},j\in \{N+1...N+M\}$ is the unlabeled 3D dataset. In our case, $M>>N$. A simple UNet can be trained with the $N$ labeled volumes to complete the segmentation task, however, with low accuracy. In our work, we aim to learn effective representation from the large amount of unlabeled data to alleviate the dependencies on labeled data for volumetric medical image segmentation.
	To this end, we propose a bootstrap representation learning method from two aspects. First, an asymmetric network with an attention-guided predictor is proposed to leverage the predictable possibilities between slices within and across different subjects for the unlabeled data. Second, a semi-supervised prototype-based foreground-background contrast mechanism is proposed to regularize the learning of the segmentation network for both labeled and unlabeled data. The overall network architecture is shown in Fig.~\ref{fig_pipeline}. Three branches are designed, one for segmentation of labeled data, and the other two branches for representation learning by leveraging the unlabeled data. Only the first branch of encoder-decoder will be used during test.
	The total objective of our method is: 
	\begin{equation}
		\mathcal{L} = \mathcal{L}_{seg} + \lambda_1 \mathcal{L}_{pred} + \lambda_2 \mathcal{L}_{fbc},
	\end{equation}
	where $L_{seg}$ is the loss function for segmentation, $L_{pred}$ is the pixel-level loss function of the predictor, and $L_{fbc}$ is the foreground/background contrastive loss. $\lambda_1, \lambda_2$ are trade-off parameters.
	
	\subsection{Representation learning by local representation prediction across and within volumes}

    We propose pixel-level prediction of feature maps between slices within and across volumes to enhance local sensitivity of the representation, which is crucial for the downstream segmentation task. The underlying assumption is spatial continuity of organs in medical volumes, which implies the representation of a slice is predictable from its neighboring slices.
    
    As shown in \emph{Representation Learning} of Fig.~\ref{fig_pipeline}, two random selected slices $x_{k_1}^{i_1}$ and $x_{k_2}^{i_2}$ from the unlabeled volumes are first processed by the encoder part $E(x)$ of the segmentation network, obtaining the local representations of the two slices $r_{k_1}^{i_1} = E(x_{k_1}^{i_1}), r_{k_2}^{i_2} = E(x_{k_2}^{i_2})$. Based on the spatial continuity of anatomical structures, an attention-guided predictor $h$ is introduced to transform the representation of one slice $r_{k_1}^{i_1}$ to predict that of the other slice $r_{k_2}^{i_2}$: $p_{k_1}^{i_1} = h(r_{k_1}^{i_1})$. 
    This prediction is optimized with a dense negative cosine similarity of each feature vector in the representation: 
    \begin{equation}
        L_{pred} = -\frac{1}{A} \sum_{i_1, i_2,k_1, k_2, u} \frac{p_{k_1}^{i_1}(u)}{\| p_{k_1}^{i_1}(u) \|_2} \cdot \frac{r_{k_2}^{i_2}(u)}{\| r_{k_2}^{i_2}(u) \|_2},
    \end{equation}
    where $u$ indicates the location in the feature map, and $A$ is a normalization factor. $i_1==i_2$ corresponds to within-volume slice prediction, while $i_1\ne i_2$ means across-volume prediction.
    
\begin{figure}[t]
		\centering
		\includegraphics[width=0.8\textwidth, trim=50 80 50 60, clip]{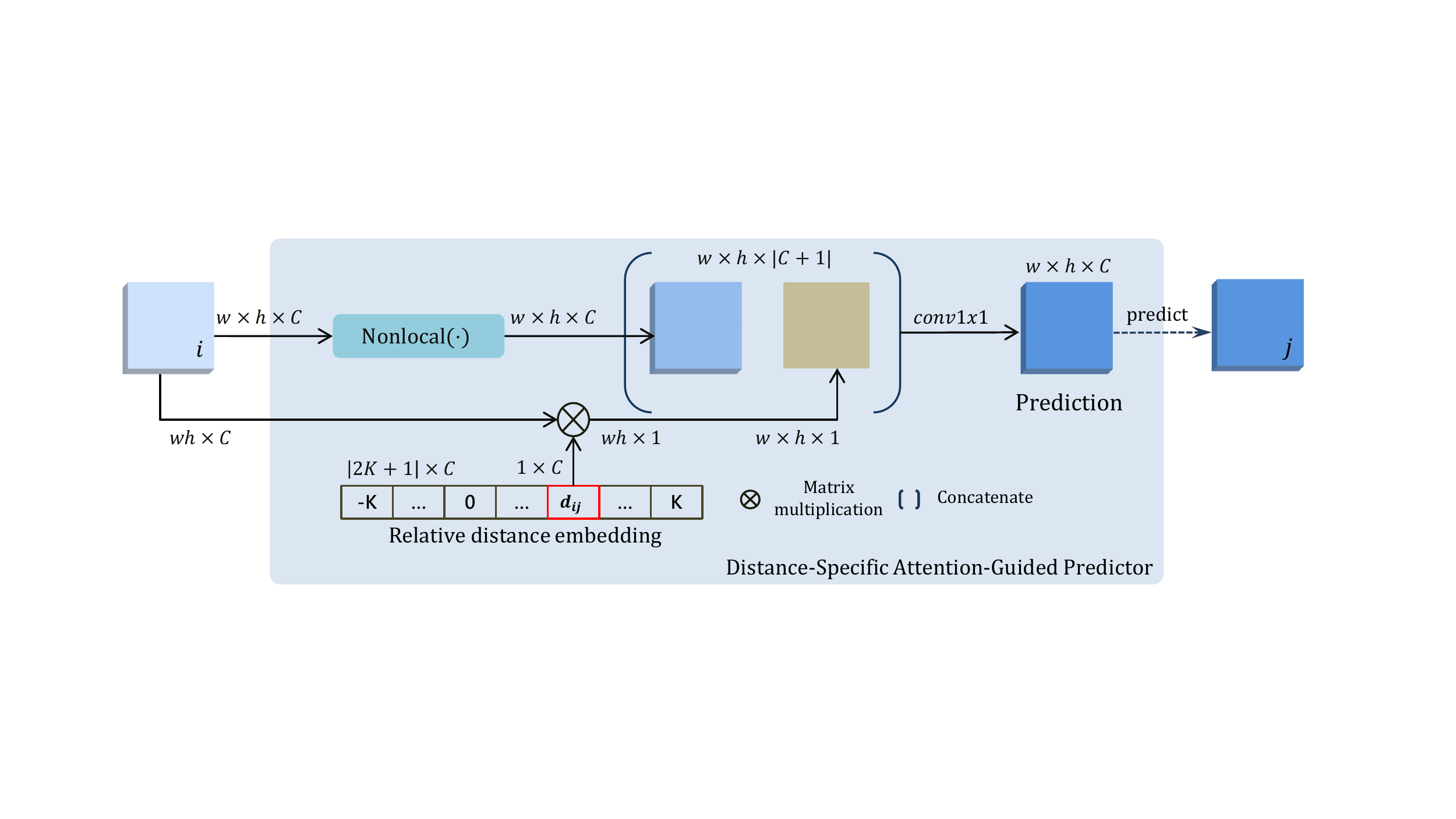}
		\caption{Detail of the distance-specific attention-guided predictor.}
		\label{fig_dap}
	\end{figure}

    \paragraph{Distance-specific attention-guided (DSAG) predictor}
    To make our dense prediction more flexible and reliable, we introduce an attention module in the predictor to leverage global context information during the prediction, and incorporate relative distance between two slices as another input of the predictor. The relative distance can somehow indicate how similar two slices can be, and therefore guide the prediction procedure.
    Fig.~\ref{fig_dap} shows the details of the proposed DSAG predictor $h$. 
    
    For the attention module, we utilize the non-local network~\cite{wang2018non} to explore the long-range dependencies in the local representations, therefore leverage the global context information which will benefit the pixel-wise representation prediction. To make the prediction capable of distinguishing target frames of different distances, a distance embedding-based feature map is concatenated with the non-local output, and then a $1\times1$ convolution layer gives the output of the predictor. 
	For the input local representation $r_{k_1}^{i_1}\in \mathcal{R}^{w\times h\times C}$ and the prediction target $r_{k_2}^{i_2}\in \mathcal{R}^{w\times h\times C}$, the embedding of relative distance $k_1-k_2$ is a vector $d\in \mathcal{R}^{C\times 1}$ selected depending on the relative distance from a learnable table of relative positional encoding~\cite{shaw2018self, ke2020rethinking}. Suppose we attempt to predict slices with maximum distance of $K$, the learnable embedding table is of length $2K+1$ and each element is a vector of dimension $C$.
    Then the distance-embedding feature map $r_d\in \mathcal{R}^{w\times h\times 1}$ can be obtained by matrix multiplication: $r_d= r_{k_1}^{i_1} \times d$. The output of the predictor is computed as follows:
	\begin{equation}
	    p_{k_2}^{i_2} = h(r_{k_1}^{i_1}) = \verb|conv|(|\verb|Nonlocal|(r_{k_1}^{i_1}), r_{k_1}^{r_1}\times d|).
	\end{equation}

	To avoid unreliable prediction between two distant slices with obvious structure difference, the relative distance of the input and the target slice is restricted by $\|k_1-k_2\|\leq d_{max}$. The interaction of relative-distance embedding (RDE) and the input representation enables a domain transfer of the RDE from a positional encoding to a feature map. 
	
    Our representation learning differs from~\cite{chen2020exploring, xie2020propagate} in three aspects. 1) Due to the spatial continuity of anatomical structures within one volume and the common spatial layout between different volumes, we use different slices within and across volumes for prediction, instead of differently augmented versions of the same image. 2) Due to the requirement of local sensitivity by the representation of the downstream segmentation task, we conduct pixel-level prediction between local feature vectors from two different slices, instead of the global vectors~\cite{chen2020exploring} or smoothed local vectors~\cite{xie2020propagate}. 3) We use a novel DSAG predictor instead of the multiple layer perception (MLP). In this way, the global context and the distance between slices can be embedded into the local representation, and make the prediction flexible and stable.
	
	\subsection{Foreground-background calibration for segmentation}
	
	While the previous representation learning part helps local feature learning of the encoder $E$ from unlabeled volumes, feature maps from the decoder $D$ directly output the segmentation results and should be more carefully explored. In this work, we propose to calibrate decoder features by enforcing consistency between labeled and unlabeled volumes, which we call foreground-background calibration (FB Calibration). Here the labeled image with ground truth provides a fair reference for the foreground in unlabeled data.
	
	The motivation of FB Calibration is that the foreground feature should be distant from its background feature, and be close to foreground features of slices from any other volumes. We use InfoNCE~\cite{oord2018representation} as the loss function to minimize violation of FB consistency between labeled and unlabeled data: 
	\begin{equation}
		\mathcal{L}_{\mathtt{InfoNCE}}(z, z^+, \{z^-\}) = 
		-\log \frac{exp(z \cdot z^+ / \tau)}{exp(z \cdot z^+ / \tau)  + \sum_{z^-} exp(z \cdot z^- / \tau)},
		\end{equation}
	where $ (z,z^+)$ makes a positive pair and $(z,z^-)$ a negative pair. $ \tau $ is a temperature hyper-parameter.
	
	As shown in \emph{Foreground-background Calibration} of Fig.~\ref{fig_pipeline}, given a slice $x_{k_1}^i$ from the labeled volume and another slice $x_{k_2}^j$ from the unlabeled volume, the intermediate feature maps from the decoder can be obtained by $z_{k_1}^i = D_l(E(x_{k_1}^i))$ and $z_{k_2}^j = D_l(E(x_{k_2}^i))$, respectively. $l$ is the index of the selected decoder layer to extract the feature maps. For FB Calibration, a crucial step is to correctly identify the foreground and background regions. 
	In our case, the foreground region of the labeled slice $x_{k_1}^i$ can be identified directly by the ground truth segmentation mask $y_{k_1}^i$. For the unlabeled slice $x_{k_2}^j$, the foreground region can be indicated by the locations with maximal response in the probability map obtained by the segmentation network, i.e., $prob(x_{k_2}^i) = \verb|softmax|(D(E(x_{k_2}^i)))$. 
	
	To avoid heavy computation and encourage stable results, for each slice, a prototype is computed by average pooling of the foreground/background region:
	\begin{equation}
	\left\{
	\begin{array}{ll}
	q_{FG}^i &= \verb|average|(z_{k_1}^i(u)| y_{k_1}^i(u)==1),\\ 
	q_{BG}^i &= \verb|average|(z_{k_1}^i(u)| y_{k_1}^i(u)==0),\\
	q_{FG}^j &= \verb|average|(z_{k_2}^j(u)| u=\verb|top-K|(prob(x_{k_2}^j)),
	\end{array}
	\right.
	\end{equation}
 	where $y_{k_1}^i$ is the annotated mask of slice $x_{k_1}^i$, $\verb|top-K|$ is an operator that return the location with top $K$ values in the probability map. It's worth noting that we do not use background prototypes for unlabeled data to avoid unstable representations caused by the diversity of structures in the background.  
 	All these prototype vectors are normalized with their length. With these foreground/background prototype vectors, FB Calibration can be implemented by optimizing the following loss function:
 	\begin{equation}
		 	\mathcal{L}_{fbc} = \frac{1}{B} \sum_{i,j} \mathcal{L}_{\mathtt{InfoNCE}}(q_{FG}^i, q_{FG}^j, \{q_{BG}^i\}),
	\end{equation}	
	where $B$ is a normalization constant.
	
	Our proposed FB Calibration has the following advantages. 1) We incorporated interaction between the labeled and unlabeled data, which can enforce the decoder features to be consistent. This differs from existing procedures~\cite{chen2020simple,chaitanya2020contrastive} that explore only unlabeled data first, and then finetune the model with labeled data. 
	2) Even for unlabeled data, we can design effective contrast local feature pairs with explicit foreground/background meaning. Existing methods~\cite{chaitanya2020contrastive} either use image-level representations, or only build contrastive pairs according to the spatial location, and without any semantic information.  
	
	\section{Experiments}
	
	\subsection{Experimental Details}
    	\paragraph{Datasets} To evaluate the proposed method, three publicly available medical datasets of volumes and sequences are used.
    	\textbf{ACDC}~\cite{bernard2018deep} consists of 100 3D short-axis cardiac cine-MRIs, captured using 1.5T and 3T scanners with expert annotations of full volume for three structures: left ventricle, myocardium, and right ventricle.
    	\textbf{Prostate}~\cite{simpson2019large} consists of 32 3D T2-weighted MRIs of the prostate region with expert annotations including two structures: peripheral zone and central gland.
    	\textbf{CAMUS}~\cite{leclerc2019deep} consists of 450 patients' 2D echocardiography sequences from end of diastole (ED) to end of systole (ES) phase with apical two-chamber (A2C) and four-chamber (A4C) views. For each sequence, three structures including the left ventricle, myocardium, and left atrium are annotated for ED and ES frames while no annotation exists for the in-between frames.
    	
    	\paragraph{Pre-processing} For fair comparison on the ACDC and Prostate datasets, we apply the same pre-processing steps as \cite{chaitanya2020contrastive}:
    	(i) intensity normalization of each 3D volume and using min-max normalization;
    	(ii) re-sampling of all 2D images and corresponding labels to a fixed resolution $ r_f $ using bi-linear and nearest-neighbour interpolation, respectively; (iii) cropping or padding with zeros to a fixed image size of $ s_f $.
    	The fixed resolutions $ r_f $ and image size $ s_f $ for each dataset are:
    	(a) ACDC: $ r_f=1.367\times1.367 mm^2 $ and $ s_f=192\times192 $,
    	(b) Prostate: $ r_f=0.625\times0.625 mm^2 $ and $ s_f=192\times192 $.
    	For the CAMUS dataset, we resample the original sequences into 10 frames and resize all the images to $ s_f=256\times256 $ and the intensity of each frame was normalized to $ [-1,1] $.
	
    	Each dataset is split into a training set $ X_{tr} $ and a test set $ X_{ts}$, consisting of volumes/sequences and their corresponding labels. 
    	For simulating the low-data regime,  $ X_{tr} $ is split into a labeled set $ D_L $, unlabeled set $ D_U $ and a validation set $ X_{vl} $.
    	The sizes of $ X_{tr} $ and $ X_{ts} $ for different datasets are:
    	(a) ACDC: $ |X_{tr}|=52 $, $ |X_{ts}|=20 $, 
    	(b) Prostate: $ |X_{tr}|=17 $, $ |X_{ts}|=15 $,
    	(c) CAMUS: $ |X_{tr}|=300 $, $ |X_{ts}|=130 $.
    	For the two volumetric datasets, $|D_L|=2,8$, and $ |X_{vl}|=2$.  
    	For the sequential dataset, we set $|D_L|=8,32$, which amounts to 16 and 64 labeled frames during training, and $ |X_{vl}|=20$.
    	The rest data in $ X_{tr}$ are used as unlabeled set $ D_U $.

    	We evaluate the performance by averaging the Dice similarity coefficient (DSC) of all structures on $ X_{ts} $ over 5 runs. 
    	$ D_L $ and $ X_{vl} $ were obtained by random sampling from $ X_{tr} $ for each run. 
    	
        \paragraph{Training details}
        In all experiments, we used the same encoder-decoder architecture as~\cite{chaitanya2020contrastive} for a fair comparison except the different backbones in~\ref{section_backbones}.
        The details of the model are described in Appendix.
        We construct each mini-batch by randomly sampling 6 slices from 2  volumes/sequences in $ D_U$, and 4 slices from $D_L$. The maximum slice distance $ d_{max} $ for each dataset is set based on the content variation of each dataset: we set the  $d_{max}$ as 8 for ACDC, 4 for Prostate, and 6 for CAMUS dataset. We used the Adam optimizer for 10000 iterations with learning rate $ 10^{-3}$.

	\subsection{Ablation study}
    In this section, we present ablation studies to investigate the effects of 1) different modules of our method, 2) selection of different predictors, 3) the prediction target of within-subject slice prediction, and 4) different backbones of the encoder.
    
        \begin{wraptable}{r}{0.4\textwidth}
	    \centering
	    \ttabbox{{}
		\caption{The effectiveness of the proposed representation learning and FB calibration with $|D_L|=2$ on ACDC.}}{%
		\label{tab_ablation_loss}
		\begin{tabular}{c|c|c|c}
			\toprule
			$ L_{seg} $ & $ L_{fbc} $ & $ L_{pred} $ & DSC \\ 
			\cline{1-4}
			\midrule			
			$ \surd $ &   		  &  	       & 0.702 \\
			$ \surd $ & $ \surd $ & 	       & 0.802 \\
			$ \surd $ &           &	$ \surd $  & 0.819 \\
			$ \surd $ & $ \surd $ &  $ \surd $ & \textbf{0.834} \\
			\bottomrule
	    \end{tabular}}
	    \end{wraptable}
        \subsubsection{Study on different modules} 
        The effectiveness of different modules of our method, i.e. our prediction module of representation learning and the FB calibration, is examined in Tab.~\ref{tab_ablation_loss}. Since only two labeled volumes are used for training, $L_{seg}$ only achieved a DSC of 0.702. Both our representation learning and FB calibration can bring the performance with a large margin, with 11.7\% and 10\% improvement of DSC, respectively.
        When the two modules are combined, even more improvement can be obtained, leading to a DSC of 0.834.
        These results can clearly show the effectiveness of the proposed representation learning and FB calibration in aspects of learning discriminative local representations.

        \begin{wraptable}{r}{0.4\textwidth}
	    \centering
	    \ttabbox{{}
		\caption{The effectiveness of the proposed DSAG predictor with $|D_L|=2$ on ACDC.}}{%
		\label{tab_ablation_predictor}
 		\begin{tabular}{c|c}
 			\toprule
 			Predictor	& DSC   \\
 			\midrule
 			MLP & 0.804 \\
 			Nonlocal & 0.821 \\
 			DSAG predictor & \textbf{0.834} \\
 			\bottomrule
 	    \end{tabular}}
	    \end{wraptable}
        \subsubsection{Selection of predictors} 
        The effectiveness of the proposed DSAG predictor is examined in Tab.~\ref{tab_ablation_predictor}. Two competitors are used: 1) MLP, which is frequently used as a predictor in previous methods~\cite{grill2020bootstrap, chen2020exploring}, and 2) Nonlocal, where the distance embedding in our module is removed. As can be seen from the table, when Nonlocal is used as the predictor, 1.7\% improvement of DSC over MLP is achieved. This indicates that the global context extracted by the Nonlocal helps a lot for the dense representation prediction. Further improvement can be obtained by the distance information embedded in the predictor, i.e., the proposed DSAG predictor, which helps distinguish slices of different positions.    
        
        \begin{wraptable}{r}{0.6\textwidth}
	    \centering
        \ttabbox{{}
        \caption{Ablation study on the task of slice prediction within subjects. Experimented with $|D_L|=2$ on ACDC and Prostate dataset, $|D_L|=8$ on CAMUS dataset.}}{%
        \label{tab_ablation_task}
        \begin{tabular}{c|c|c|c|c}
            \toprule
            \multirow{2}{*}{Slice Prediction} & \multirow{2}{*}{ACDC} & \multirow{2}{*}{Prostate} & \multicolumn{2}{c}{CAMUS} \\
              & & & A2C & A4C \\
            \midrule
            RE                 & 0.814 & \textbf{0.627}    & 0.737 & 0.749 \\
            IR-NL & 0.821 & 0.627    & 0.742 & 0.766 \\
            IR-DSAG   & \textbf{0.827} & 0.626    & \textbf{0.766} & \textbf{0.778} \\
            \bottomrule
        \end{tabular}}
	    \end{wraptable}
        \subsubsection{Target of within-subject slice prediction: distance-specific or distance-agnostic}
        We compared three different prediction targets for within-subject slice prediction: 1) the representation expectation (\textbf{RE}) of all the slices within a volume, which is a simple and straightforward baseline when we do the distance-agnostic prediction, 2) the individual representation of each slice, with no distance reference, based on only the distance-agnostic non-local module in the predictor (\textbf{IR-NL}), and 3) individual representation of each slice, but with distance reference, based on our DSAG predictor (\textbf{IR-DSAG}). In the above three tasks, the first and second ones are based on the plain NonLocal module without distance embedding.
    
        In the scenario of distance-agnostic prediction, predicting expectation is very straightforward. Compared with the individual representation prediction, i.e. \textbf{IR-NL} in Tab.~\ref{tab_ablation_task}, the simple baseline of \textbf{RE} performs 0.7\% worse on ACDC and 1.7\% worse on A4C of CAMUS. It may be because \textbf{IR-NL} is a more challenging task that can better facilitate representation mining. Without distance-embedding, however, the predictor is blind to know how many changes can happen. To this end, we introduced the DSAG predictor, to provide distance embedding and let the predictor better guess the changes within a relative distance. The experimental results also proved this point. As can be seen in Tab.~\ref{tab_ablation_task}, our \textbf{IR-DSAG} achieves the best results compared to the distance-agnostic ones. It surpasses the baseline with 1.3\% and 2.9\% DSC on ACDC and CAMUS datasets, respectively. For the Prostate dataset, all three tasks perform similarly, which can be explained by the reason that the slices change more sharply on the Prostate dataset than on the other two datasets. Distant slices may share a low similarity, leading to increasing difficulty of predicting changes in specific distance and less effectiveness of introducing additional distance embedding. 
        
        \begin{wraptable}{r}{0.65\textwidth}
	    \centering
        \ttabbox{{}
        \caption{Comparison of proposed method with CGL for different backbones of the encoder. Experimented on ACDC dataset.}}{%
        \label{tab_ablation_backbone}
            \begin{tabular}{c|c|c|c|c}
            \toprule
            \multirow{2}{*}{Method} & \multicolumn{2}{c|}{VGG13} & \multicolumn{2}{c}{ResNet18} \\ 
             & $|D_L|=2$ & $|D_L|=8$ & $|D_L|=2$ & $|D_L|=8$  \\
            \midrule
            Rand. Init.                        & 0.713 & 0.850 & 0.723 & 0.854 \\
            CGL~\cite{chaitanya2020contrastive} & 0.771 & 0.872 & 0.784 & 0.870 \\
            Ours                                & \textbf{0.816} & \textbf{0.888} & \textbf{0.823} & \textbf{0.884} \\
        \bottomrule
        \end{tabular}}
        \end{wraptable}
        \subsubsection{Generalization to other backbones}
        \label{section_backbones}
        We investigated the effectiveness of the proposed method with two well-known backbones for the encoder: 1) VGG13~\cite{simonyan2015very} and ResNet18~\cite{he2016deep}. To adapt ResNet to the U-shape architecture with skip connection, we replaced the first $7\times7$ convolution with two $3\times3$ convolutions followed by a $3\times3$ maxpooling with a stride of 2. This adaptation enables a skip connection of finer low-level representation to the decoder. We set the output channel dimension of the projection head as 512 on both the backbones of VGG13 and ResNet18 for all the methods. As can be seen in Tab.~\ref{tab_ablation_backbone}, the proposed method works the best across different backbones. Significant performance improvements are achieved by the proposed representation learning and FB calibration. These results prove the superiority and generalization of our method. 

    	\begin{table}[pt]
    	   \centering
    		\ttabbox{\caption{Comparison of the proposed method with other existing methods on ACDC and Prostate.}}{%
        	\label{compare_methods}
    		\begin{tabular}{c|c|c|c|c}
    			\toprule
    			& \multicolumn{2}{c|}{ACDC} & \multicolumn{2}{c}{Prostate}\\
    			Methods & $ |D_L|=2 $ & $ |D_L|=8 $ & $ |D_L|=2 $ & $ |D_L|=8 $ \\	\cline{1-5}		
    			\midrule			
    			Random init. 				 	  	& 0.702 & 0.844 & 0.550 & 0.636 \\
    			Self-train\cite{bai2017semi} 	  	& 0.749 & 0.860 & 0.598 & 0.680 \\
    			MixUp Baseline\cite{zhang2017mixup}   	  	& 0.785 & 0.863 & 0.593 & 0.661 \\
    			Data Aug.\cite{chaitanya2021semi} 	& 0.786 & 0.865 & 0.597 & 0.667 \\
    			CGL\cite{chaitanya2020contrastive} & 0.789 & 0.872 & 0.619 & 0.684 \\
    			Ours						    & \textbf{0.834} & \textbf{0.892} & \textbf{0.636} & \textbf{0.690}   \\
    			\midrule
    			Benchmark & \multicolumn{2}{c|}{$ (|D_L|=78) $ 0.912} & \multicolumn{2}{c}{$ (|D_L|=20) $ 0.697} \\		
    			\bottomrule
    		\end{tabular}}
    	\end{table}
    	\begin{table}[pt]
    		\ttabbox{{}\caption{Comparison of the proposed method with other existing methods on CAMUS.}}{%
    		\label{compare_camus}
    		\centering
    		\begin{tabular}{c|c|c|c|c}
    			\toprule
    			& \multicolumn{2}{c|}{A2C} & \multicolumn{2}{c}{A4C}\\
    			Methods & $ |D_L|=8 $ & $ |D_L|=32 $ & $ |D_L|=8 $ & $ |D_L|=32 $ \\	\cline{1-5}		
    			\midrule			
    			Random init. 				 	  	& 0.695 & 0.819 & 0.714 & 0.793 \\
    			CGL.\cite{chaitanya2020contrastive} & 0.730 & 0.829 & 0.764 & 0.836 \\	
    			MixUp Baseline\cite{zhang2017mixup}   	  	& 0.745 & 0.840 & 0.775 & 0.838 \\					
    			Ours	& \textbf{0.768} & \textbf{0.846} & \textbf{0.786} & \textbf{0.845} \\
    			\midrule
    			Benchmark & \multicolumn{2}{c|}{$ (|D_L|=300) $ 0.898} & \multicolumn{2}{c}{$ (|D_L|=300) $ 0.905} \\
    			\bottomrule
    		\end{tabular}}
    	\end{table}

        \subsection{Comparison with state-of-the-art methods} 
       
        \paragraph{Volumetric datasets of ACDC and Prostate}
        We evaluate on two benchmark volumetric datasets, i.e. ACDC and Prostate, on two settings with different annotated volumes respectively. 
        To provide a comprehensive comparison, here we also provide a baseline by extending a popular semi-supervised learning method~\cite{zhang2017mixup} to segmentation by altering classification loss to the segmentation dice loss. 
        Obviously from the Tab.~\ref{compare_methods} our experimental results proved our superiority over previous benchmarks~\cite{chaitanya2020contrastive}. Our method outperforms the previous state-of-the-art, i.e. CGL~\cite{chaitanya2020contrastive}, with a large margin of 4.5\% DSC on ACDC using 2 volumes and 2\% DSC using 8 volumes. For the more challenging Prostate Dataset, our bootstrap mining method can capture its inherent attributes and enforce better representation. Our model surpasses the CGL with 1.7\% averaged Dice using 2 volumes. This attests that mining the representation by exploring the predictability based on the spatial continuity of anatomical structures is more suited to the volumetric and sequential data than contrastive methods. This could because continuity can benefit prediction, but not contrast. Worthy to note that, the gap between our performance of using 8 volumes and a benchmark-setting of using 20 volumes has been narrowed to 0.7\% averaged Dice. 
    	
        \paragraph{Sequential dataset of CAMUS}
        To comparison on the sequential dataset of CAMUS, we construct a baseline of supervised learning on the same labeled training set with ours from random initialization, denoted as \textbf{Random init.} in Tab.~\ref{compare_camus}. In addition, we also provide two strong baselines based on MixUp and CGL respectively, refer to our Appendix for details. We evaluate our method on the sequences of two views on the CAMUS dataset. From the table, we can see that our method outperforms the strong baselines of MixUp on all settings. In addition, we only have a gap of 5.2\% DSC but saving 89.3\% annotations. This sufficiently proved the effectiveness of our method on the sequential data.

	\section{Conclusion}
    
    In this paper, we presented a method of bootstrap representation learning for the segmentation of medical volumes and sequences. Specifically, based on the spatial continuity of neighboring slices/frames and the common spatial layout of anatomical structures, we proposed a within-and-across volumes slice prediction mechanism for representation learning from the unlabeled data, and a foreground-background calibration module for feature map calibration of unlabeled data with references features from labeled data. Experiments on three benchmark datasets demonstrated that our proposed method can effectively improve the segmentation performance when only a few labeled data are available. Due to the underlying assumption for spatial/temporal continuity, our proposed method may need further adjustment for the segmentation of irregular lesions in medical data. We will leave this for our future work. 
  
	{
		\small
		\bibliographystyle{plain}
		\bibliography{references}
	}
	
	\newpage
	\appendix
	\section{Appendix}
	
	In this appendix, we add more details of our work, including more ablation study, parameter selection, network architecture, and visualization results.
	The ablation studies provide more insights about our method:
	
	\begin{itemize}
	\item We provided results by incrementally adding the relationship within volumes and among volumes on the basic model, to attest to the effectiveness of our within- and across-subject representation mining strategy, respectively. 
	
	\item We studied the selection of the decoder layer for FB calibration and proved that calibration with a middle-level representation, where each location represents a small local region of the original image, achieved the best result.
	\end{itemize}
		
	\subsection{Training details}
    \textbf{Data splitting}: Training set $ X_{tr} $ and test set $ X_{ts} $ were chosen according to the code~\footnote[1]{\url{https://github.com/krishnabits001/domain_specific_cl}} of~\cite{chaitanya2020contrastive} to have a fair study.
    
    \textbf{Data augmentation}: In our implementation, we use the same data augmentation as in~\cite{chaitanya2020contrastive} across all methods, including rotation, flipping, Gaussian blur, contrast, and brightness changes. For the unlabeled data $D_U$ which is involved in the slice prediction, we treat the volume as a whole and apply the same augmentation to all the slices in that volume. 

    \textbf{Loss function}: The loss function for segmentation $L_{seg}$ is the dice loss~\cite{milletari2016v}, which is widely used for medical image segmentation tasks and can alleviate the issue of unbalanced foreground and background region. The trade-off parameters of $\lambda$ for the other two loss function $L_{pred}$ and $L_{fbc}$ are set as 0.5 and 0.5, separately, according to the performance on a validation set. For the InforNCE loss (Eq.4), we use the temperature parameter $\tau=0.1$ by adopting from~\cite{chaitanya2020contrastive}. 
    We use the cosine annealing~\cite{loshchilov10sgdr} strategy to construct a learning rate schedule. 
    
    \subsection{Configuration of two competitors}
    \textbf{CGL}: In our experiment, the results of CGL on datasets of ACDC and Prostate are obtained from the original publication. To train a strong baseline on the sequential dataset CAMUS, our implementation of CGL~\cite{chaitanya2020contrastive} and the training procedure closely matches that of~\cite{chaitanya2020contrastive}, except for the following differences.
    Instead of $(G^D + L^D)$~\footnote{$G^D$, the positive set is obtained by images from corresponding partitions across volumes. The negative set is obtained by images only from other partitions. $L^D$, the positive set is obtained by representations from corresponding local regions across volumes. The negative set is obtained by representations from the remaining local regions.}, we use the random sampling strategy $(G^R + L^R)$, which only takes augmentation from the same slice as positive. Because we empirically found that the $ L^R $ worked better than $ L^D $ and inferred that the rough partition and alignment may not provide stable correspondence on sequential data of CAMUS. \cite{chaitanya2020contrastive} also proved that, in their framework, $ (G^R + L^R)$ performs better for scenarios of large temporal changes, in their experiments on Cityscapes Dataset~\cite{cordts2016cityscapes}. 

    \textbf{MixUp baseline}: To extend Mixup~\cite{zhang2017mixup} to the task of segmentation for medical images, we replace the cross-entropy loss to the dice loss~\cite{milletari2016v}. The feature-target pairs are obtained by randomly combining two images from the current batch, and the strength of interpolation is controlled by the hyper-parameter $\alpha$~\cite{zhang2017mixup}. We experimented with 3 values of $\alpha$: 0.1, 0.2 and 0.3. We observed that these different values gave similar performances on the validation set. We set it as 0.1 for stability consideration on the ultrasound image.

    \subsection{More ablation studies}
    
    Here, we present more ablation studies to investigate the effects of 1) slice prediction within- and across subjects in the representation mining, and 4) the selection of the decoder layer in FB calibration.

    \subsubsection{Within- and across-subject slice prediction}
    \begin{wraptable}{r}{0.6\textwidth}
        \centering
        \ttabbox{{}\caption{Ablation study on the representation mining of slice prediction within- and across-subjects. $|D_L|=2$ for ACDC and Prostate datasets, and $|D_L|=8$ for CAMUS.}}{%
        \label{tab_ablation_across_within}
        \begin{tabular}{c|c|c|c|c|c}
            \toprule
            \multirow{2}{*}{across} & \multirow{2}{*}{within} & \multirow{2}{*}{ACDC} & \multirow{2}{*}{Prostate} & \multicolumn{2}{c}{CAMUS} \\
                    &         &       &          & A2C   & A4C   \\
            \midrule
            $\surd$ &         & 0.817 & 0.619    & 0.751 & 0.774 \\
                    & $\surd$ & 0.827 & 0.626    & 0.766 & 0.778 \\
            $\surd$ & $\surd$ & \textbf{0.834} & \textbf{0.636}    & \textbf{0.768} & \textbf{0.786} \\
            \bottomrule
        \end{tabular}}
    \end{wraptable}
    We investigated the effect of the representation mining strategy, i.e., the predictable relationship within-subject and across-subjects, in Tab.~\ref{tab_ablation_across_within}. Both kinds of relationships can benefit our model. We interestingly observed that mining slice predictability within-subject yield more gains than across-subjects for all three datasets. 
    This is probably because of the inherent difference among subjects, such as the start point of the slices, slice intervals, and coverage ranges.
    Moreover, the combination of slice prediction within- and cross-subjects further boost the accuracy by about 1\% DSC, indicating the complementary of the two types of predictability.
    
	\begin{wraptable}{r}{0.6\textwidth}
	    \centering
	    \ttabbox{{}
		\caption{Selection of the decoder layer for FB calibration, experimented with $|D_L|=2$ on ACDC dataset.}}{%
		\label{ablation_declayer}
    		\begin{tabular}{c|c|c|c|c}
    			\toprule
    			layer $ l $	& layer 1 & layer 2 & layer 3 & layer 4 \\
    			\midrule
    			ACDC & 0.805 & 0.818 & \textbf{0.834} & 0.806 \\
    			\bottomrule
    		\end{tabular}}
    \end{wraptable}
    \subsubsection{Selection of the decoder layer for FB calibration}
    We investigate the effect of different decoder layers for FB Calibration. Feature maps extracted from layer $l=\{1,2,3,4\}$ are examined. From Tab.~\ref{ablation_declayer}, we can observe that the best performance is obtained when layer 3 is used. This indicates that the calibration works better on the intermediate level with a suitable resolution. Due to the inaccurate foreground/background identification for unlabeled volumes, FB calibration with feature maps of high resolution in the early stage of learning may lead to unstable training.

    \subsection{Network Architecture}
    The details of our baseline network UNet~\cite{ronneberger2015u} in training, including an encoder, a predictor, and a decoder, are shown in Tab.~\ref{table_unet}. During inference, the predictor will be discarded. The encoder consists of 6 layers, including convolution blocks (ConvBlock) and maxpooling. Each ConvBlock consists of a $3\times3$ convolution followed by ReLU activation and batch normalization. The decoder consists of 6 layers, including ConvBlocks, upsampling using nearest neighbor. Skip connections are deployed between corresponding layers of the encoder and decoder. 

    \begin{table}[h]
    	\centering
    	\ttabbox{{}\caption{The configuration of the network architecture in training. The predictor is discarded in inference.}}{%
    	\label{table_unet}
    	\begin{tabular}{c|l|c|c}
    	    \toprule
    		Layer & Architecture & Channels & Resolution \\
    		\cline{1-4} 
    		\multicolumn{4}{c}{Encoder} \\
    		\midrule
    		1 & ConvBlock*2  & 16 & $ 192 \times 192 $ \\
    		2 & Maxpool(2,2) + ConvBlock*2 &  32 & $ 96 \times 96 $ \\
    		3 & Maxpool(2,2) + ConvBlock*2 &  64 & $ 48 \times 96 $ \\
    		4 & Maxpool(2,2) + ConvBlock*2 & 128 & $ 24 \times 24 $ \\
    		5 & Maxpool(2,2) + ConvBlock*2 & 128 & $ 12 \times 12 $ \\
    		6 & Maxpool(2,2) + ConvBlock*2 & 128 & $ 6 \times 6 $ \\
    		\midrule 
    		\multicolumn{4}{c}{Predictor} \\
    		\midrule
    		  & DSAG predictor  & 128 & $6\times6$ \\
    		\midrule
    		\multicolumn{4}{c}{Decoder} \\
    		\midrule
    		1 & Upsample + ConvBlock + Cat(enc5, dec1) + ConvBlock*2  &  16 & $ 12 \times 12 $ \\
    		2 & Upsample + ConvBlock + Cat(enc4, dec2) + ConvBlock*2  &  32 & $ 24 \times 24 $ \\
    		3 & Upsample + ConvBlock + Cat(enc3, dec3) + ConvBlock*2  &  64 & $ 48 \times 48 $ \\
    		4 & Upsample + ConvBlock + Cat(enc2, dec4) + ConvBlock*2  & 128 & $ 96 \times 96 $ \\
    		5 & Upsample + ConvBlock + Cat(enc1, dec5) + ConvBlock*2  & 128 & $ 192 \times 192 $ \\
    		\midrule 
    		\multicolumn{4}{c}{Segmentation layer} \\
    		\midrule
    		  & ConvBlock*1 + Conv3x3 & classes & $ 192 \times 192 $ \\
    		\bottomrule
    	\end{tabular}}
    \end{table}

    \begin{figure}[t]
        \centering
        \includegraphics[width=\textwidth]{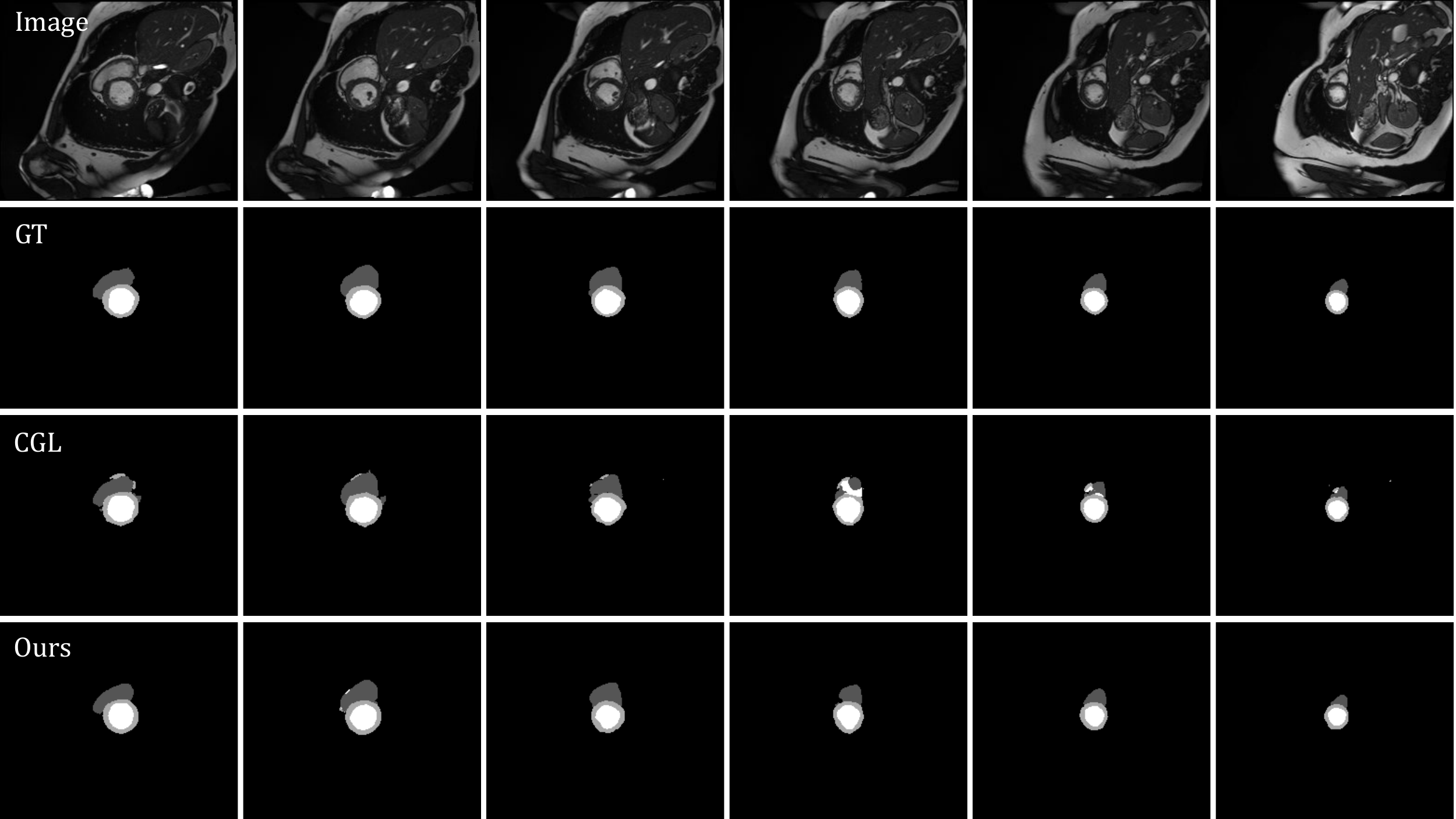}
        \caption{Visualization of segmentation results on ACDC with $|D_L|=2$. From top to bottom: image, ground truth, results from CGL~\cite{chaitanya2020contrastive} and our method.}
        \label{fig_acdc_seg}
    \end{figure}
    \begin{figure}[t]
        \centering
        \includegraphics[width=0.8\textwidth, trim=0 0 200 0, clip]{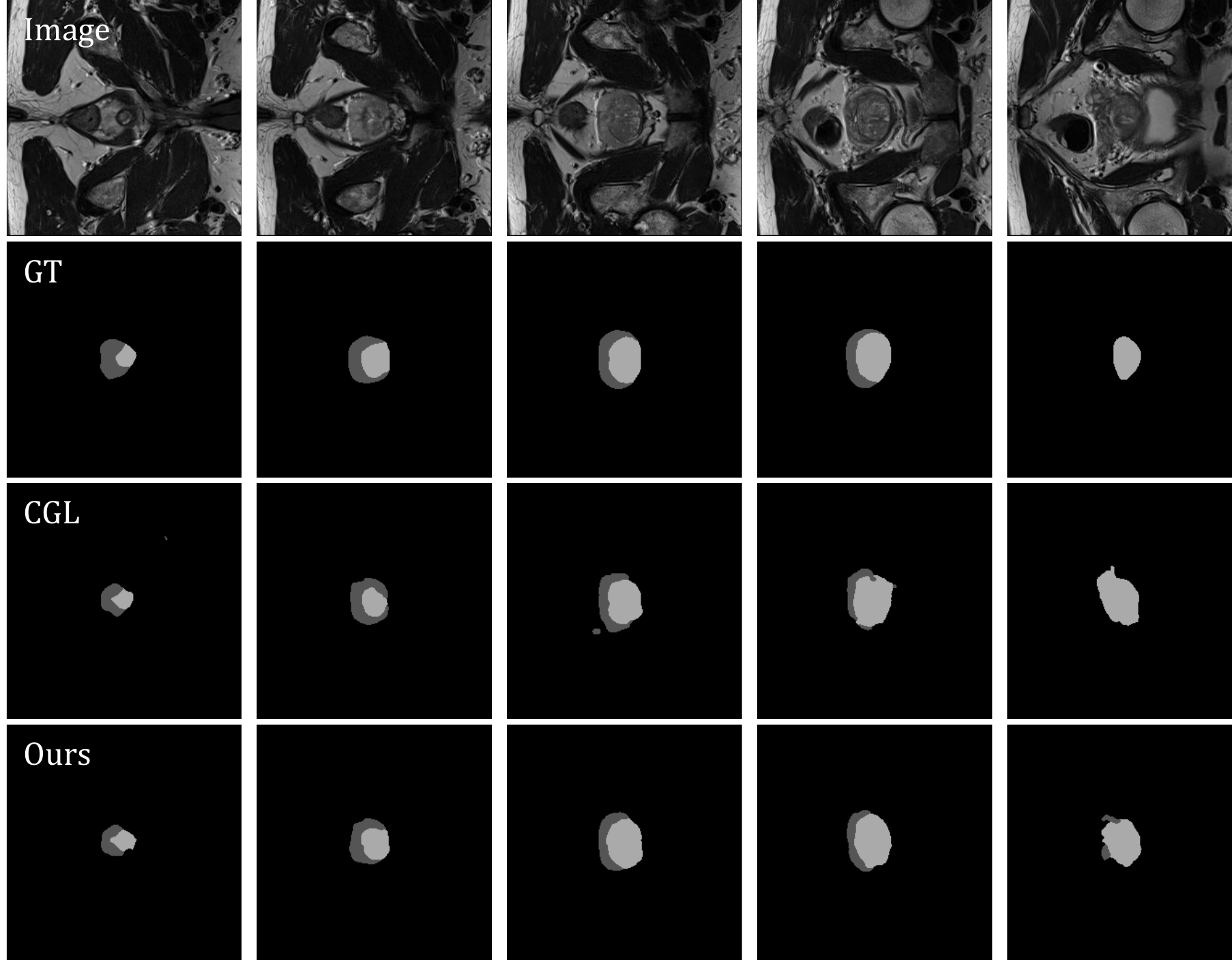}
        \caption{Visualization of segmentation results on Prostate with $|D_L|=2$. From top to bottom: image, ground truth, results from CGL~\cite{chaitanya2020contrastive} and our method.}
        \label{fig_prostate_seg}
    \end{figure}
    \begin{figure}[t]
        \centering
        \includegraphics[width=0.8\textwidth, trim=0 0 80 0, clip]{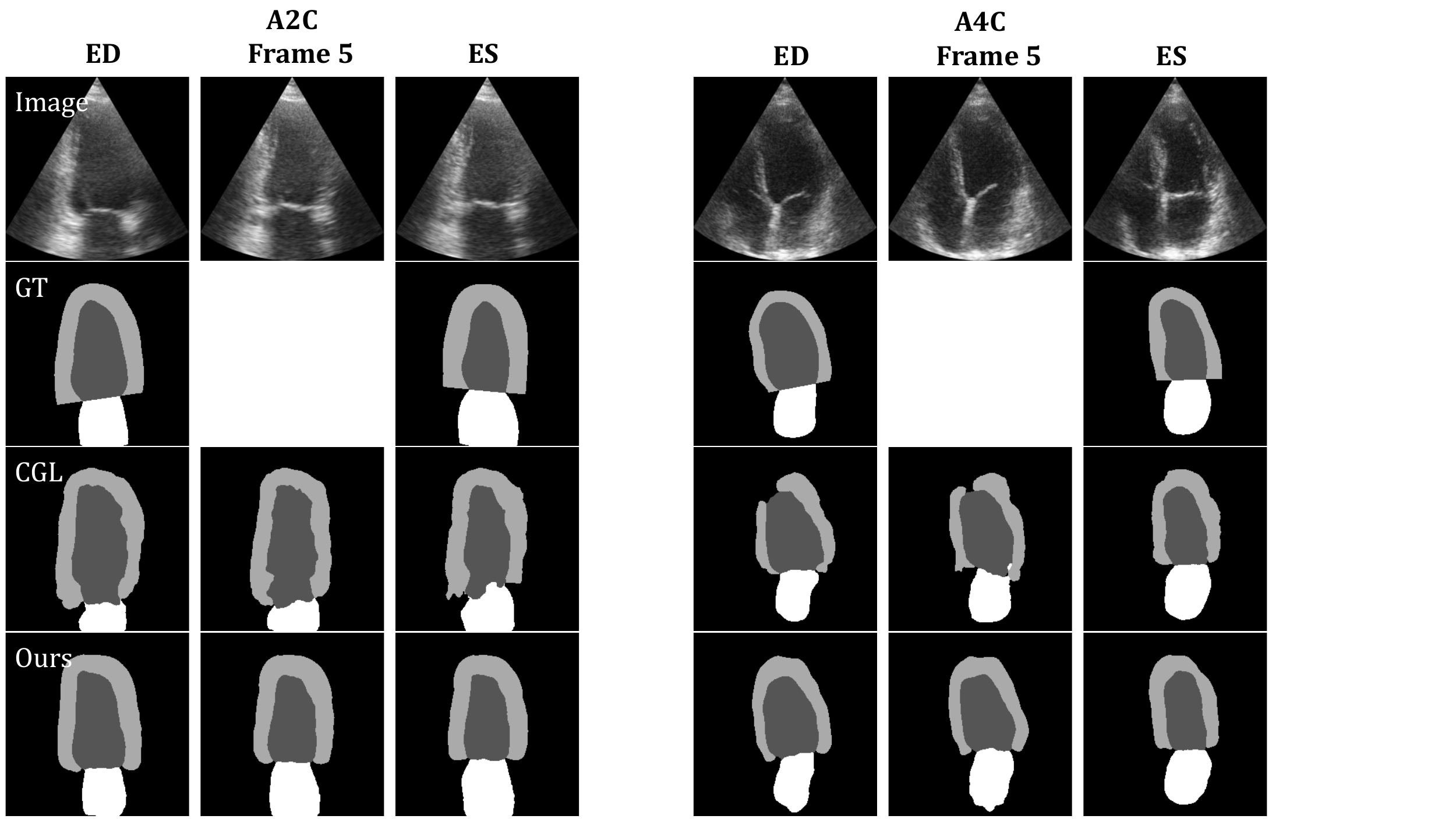}
        \caption{Visualization of segmentation results on CAMUS with $|D_L|=8$. From top to bottom: image, ground truth (GT), results from CGL~\cite{chaitanya2020contrastive} and our method. The left three columns show that the results from A2C sequence and the right from the A4C sequeunce. For each sequence, three frames are demonstrated, including the ED, ES frames and one unlabeled in-between frame.}
        \label{fig_camus_seg}
    \end{figure}
    
    \subsection{Visualization of segmentation results}
    We visualized the segmentation examples of our method and CGL~\cite{chaitanya2020contrastive} on the three datasets in Figs.~\ref{fig_acdc_seg},~\ref{fig_prostate_seg},~\ref{fig_camus_seg}.
    We can see that although only two labeled volumes were used during training, our method can predict pretty well the structures of cardiac and prostate of difference slices, and better than that of its competitor CGL. For the sequential echocardiography data of CAMUS, where only two frames were annotated for each sequence, our method can give good segmentation for the ED/ES frames and the in-between frames with only 8 sequences used for training. Compared to CGL, our method gives smoother boundaries without missing parts. Also, the continuity of the segmented structures between frames is better preserved by our method than by CGL, indicating the effectiveness of our slice prediction-based representation mining and FB calibration. 
    
\end{document}